\documentclass{article}

\usepackage{microtype}
\usepackage{graphicx}
\usepackage{subfigure}
\usepackage{booktabs} 

\usepackage{amsmath,amssymb,dsfont} 
\usepackage{mathtools}

\usepackage[accepted]{icml2019}

\icmltitlerunning{Learning latent state representation for speeding up exploration}

\begin{document}

\twocolumn[
\icmltitle{Learning latent state representation for speeding up exploration}



\icmlsetsymbol{equal}{*}

\begin{icmlauthorlist}
\icmlauthor{Giulia Vezzani}{iit}
\icmlauthor{Abhishek Gupta}{ucb}
\icmlauthor{Lorenzo Natale}{iit}
\icmlauthor{Pieter Abbeel}{ucb,cov}

\end{icmlauthorlist}

\icmlaffiliation{iit}{Istituto Italiano di Tecnologia, Genoa, Italy.}
\icmlaffiliation{ucb}{University of California Berkely, California.}
\icmlaffiliation{cov}{Covariant.ai,  Emeryville, California}

\icmlcorrespondingauthor{Giulia Vezzani}{giulia.vezzani@iit.it}

\icmlkeywords{Exploration in RL, Representation Learning, ICML workshop}

\vskip 0.3in
]



\printAffiliationsAndNotice{}  

\begin{abstract}
Exploration is an extremely challenging problem in reinforcement learning, especially in high dimensional state and action spaces and when only sparse rewards are available. Effective representations can indicate which components of the state are task relevant and thus reduce the dimensionality of the space to explore. In this work, we take a representation learning viewpoint on exploration, utilizing prior experience to learn effective latent representations, which can subsequently indicate which regions to explore. Prior experience on separate but related tasks help learn representations of the state which are effective at predicting instantaneous rewards. These learned representations can then be used with an entropy-based exploration method to effectively perform exploration in high dimensional spaces by effectively lowering the dimensionality of the search space. We show the benefits of this representation for meta-exploration  in a simulated object pushing environment.
\end{abstract}

\section{Introduction}
\label{intro}
Efficiently exploring the state space to fast experience the reward is a crucial problem in Reinforcement Learning (RL) and assumes even more relevance when dealing with real world tasks, where usually the only obtainable reward functions are \textit{sparse}. In recent years, several diverse strategies have been developed to address the exploration problem~\cite{chentanez2005intrinsically, stadie2015incentivizing, bellemare2016unifying, plappert2017parameter, fortunato2017noisy, van2015learning, sendonaris2017learning, vecerik2017leveraging, taylor2011integrating, nair2018overcoming, rss}.
Most works assume that exploration is performed with no prior knowledge about the solution of the task.  This assumption is not necessarily realistic and surely does not hold for humans that constantly use their knowledge and past experience to solve new tasks. 
The idea of exploiting knowledge from prior tasks to fast adapt to the solution of a new task is a new promising approach already widely used in Deep RL \cite{finn2017model, clavera2018model}, and some works have specifically considered using this for addressing the exploration problem~\cite{gupta2018meta}.

In this work, we leverage on prior experience to learn a latent representation  encoding the components of the state that are task relevant and thus reduce the dimensionality of the space to explore. In particular, we use the solutions of separate but related tasks to learn a minimal shared representation effective at predicting the rewards. Such a representation is then  used during the solution of new tasks to focus exploration only in a sub-region of the whole state space, which, in its entirety, might instead contain irrelevant factors that would make the search space especially large.

The paper is organized as follows. Section \ref{sota} briefly reviews some relevant work on the exploration problem in RL. After the definition of the mathematical notation used (Section \ref{background}), we describe the proposed algorithm in Section \ref{method}. Sections \ref{results} and \ref{conclusions} ends the paper showing some relevant results and discussion about the approach. More information on the algorithm and further experiments are collected in the Appendix.

\section{Related work}
\label{sota}
 Several exploration strategies have been proposed in the last years comprising different  criteria used for encouraging exploration. In~\cite{chentanez2005intrinsically, stadie2015incentivizing}, the exploration is based on \textit{intrinsic motivation}. During the training, the agent learns also a model of the system and  an \textit{exploration bonus} is assigned when novel states with respect to the trained  model are encountered. Novel states are identified as those states that create a stronger disagreement with the model trained until that moment. Another group of exploration algorithms are \textit{count-based methods} that directly count the number of times a certain state has been visited to guide the agent towards states less visited. Obviously, such an approach is infeasible in continuous state space. For this reason, some works such as~\cite{bellemare2016unifying} extend count-based exploration approaches to non-tabular (continous) RL using density models to derive a pseudo-count of the visited states.
Another approach to encourage exploration consists of injecting noise to the agent's parameters, leading to richer set of agent behaviors during training~\cite{plappert2017parameter, fortunato2017noisy}. 
These exploration strategies are task agnostic in that they aim at providing good exploration without exploiting any specific information of the task itself. More recently instead, the exploration problem has been cast into \textit{meta-learning} (or learning to learn), the field of machine learning whose goal is to learn strategies for fast adaptation by using prior tasks~\cite{finn2017model}. An example of application of meta-learning for the exploration problem is shown in~\cite{gupta2018meta}, where a novel algorithm is presented to learn exploration strategies from prior experience.  
Alternatively, it is possible to get around the exploration problem by providing task demonstrations for guiding and speeding up the training~\cite{van2015learning, sendonaris2017learning, vecerik2017leveraging, taylor2011integrating, nair2018overcoming}. The work presented in~\cite{rss} shows how the proper incorporation of human demonstrations into RL methods allows reducing the number of samples required for training an agent to solve complex dexterous manipulation tasks with a multi-fingered hand. 

The algorithm we present in this work  aims at combining the benefits of some of the most popular exploration approaches. In particular, similarly to extended count-based methods, our method encourages exploration in portions of the states not visited yet   but, instead of searching in the entire state space, exploration is focused on those regions learnt to be task relevant from prior experience, as suggested by  \textit{meta-learning}.

\section{Preliminaries}
\label{background}
We model the control problem as a Markov decision process (MDP), which is defined using the tuple: \hbox{$\mathcal{M}=\{\mathcal{S}, \mathcal{A}, \mathcal{P}_{sa}, R, \gamma, \rho_0 \}$}. $\mathcal{S} \subseteq \mathbb{R}^n$ and $\mathcal{A} \subseteq \mathbb{R}^m$ represent the state and actions. \hbox{$R : \mathcal{S} \times \mathcal{A} \rightarrow \mathbb{R}$} is the reward function which measures task progress and is considered to be \textit{sparse} in this context. $\mathcal{P}_{sa} : \mathcal{S} \times \mathcal{A} \rightarrow \mathcal{S}$ is the transition dynamics, which can be stochastic. In model-free RL, we do not assume knowledge about this transition function, and require only sampling access to this function. $\rho_0$ is the probability distribution over initial states and $\gamma \in [0,1)$ is a discount factor.  We wish to solve for a stochastic policy of the form \hbox{$\pi: \mathcal{S} \rightarrow \mathcal{A}$}, which optimizes the expected sum of rewards:
\begin{equation}
\label{goal}
\eta(\pi) = E_{\pi, \mathcal{M}} \Bigg[ \sum_{t=0}^\infty \gamma^t R_t \Bigg].
\end{equation}
\section{Method}
\label{method}
We consider a family of tasks sampled from a distribution $P_{T}$.
We  assume $N$ tasks $T_1, \dots, T_N \sim P_{T}$ to be already solved,  and to have access to  the states visited during each training $\{S^{(i)}\}_{i=1}^N$ together with the  experienced rewards signals $\{R^{(i)}\}_{i=1}^N$. The set of states  $S^{(i)}$  of the $i$-th task includes all the trajectories $\tau_t^{(i)}$ experienced during each training step $t$:
 	$S^{(i)}=\{\tau_t^{(i)}\}_{t=1}^{t_{max}}$,
 	  with $t_{max}$ the number of training steps, $\tau=\{s_0 \in \mathbb{R}^n, \dots, s_l \in \mathbb{R}^n\}$ and $l$ the length of each trajectory. Analogously for the rewards: $
 	  R^{(i)}=\{r_t^{(i)}\}_{t=1}^{t_{max}}$,
 	   with $t_{max}$ the number of training steps, $r=\{R_0 \in \mathbb{R}, \dots, R_l \in \mathbb{R}\}$ and $l$ the length of each trajectory.
 The goal is to efficiently use the information that can be extracted from the $N$ tasks for fast exploration during the solution of a new task $T_{N+1} \sim P_{T}$.

The key idea of this work consists of  learning from the $N$ prior tasks a latent representation $z \in \mathbb{R}^p$ ($p<n$) of that portion of the state mostly affecting the experience of rewards. During the solution of the new  task $T_{N+1}$, the latent representation $z $ is then considered the subregion of the state on which to focus the exploration.  

At this aim, we design 1) a \textit{multi-headed framework} for reward regression on the 
	tasks $T_1, \dots, T_N \sim P_{T}$ to encode in the  shared layers of the network the latent representation $z $; 2) a \textit{novel exploration strategy} consisting of the maximization of the entropy over the latent representation $z$ rather than over the entire state space.


\subsection{Learning latent state representation from prior states using a multi-headed network}
\label{multihead}
We assume $N$ tasks to be sampled from the same distribution $P_{T}$. An example of task distribution is given by the object-pusher task (Fig. \ref{fig:pusher-env}): a manipulator is required to push one specific object (among other objects) towards a target position.
 Each task  differs in the initial objects and goal positions. Our intuition suggests   that solving tasks sampled from the same distribution $P_{T}$ must provide some information useful
for the solution of a new task $T_{N+1} \sim P_{T}$.   In particular, we aim at using past experience to learn the portion of the state that is relevant to be explored for fast experiencing the rewards. Focusing exploration only on a subregion of the space is in fact essential when dealing with large state space or very sparse reward functions.
In our example, we can easily notice that moving the object of interest is what really matters for the task solution, regardless of the other objects positions. Although it might be easy sometimes to  derive similar arguments, inferring a proper latent representation is not  straightforward in general, especially for harder tasks.   A possible way to extract this information  is to learn the state features that are most predictive of the reward signals collected during the solution of  past tasks.  A minimal state representation that is predictive of the reward represents in fact the region the agent needs to explore to faster experience the rewards and, consequently, solve a new task. 

\begin{figure}
\centering
	\includegraphics[width=0.7\columnwidth]{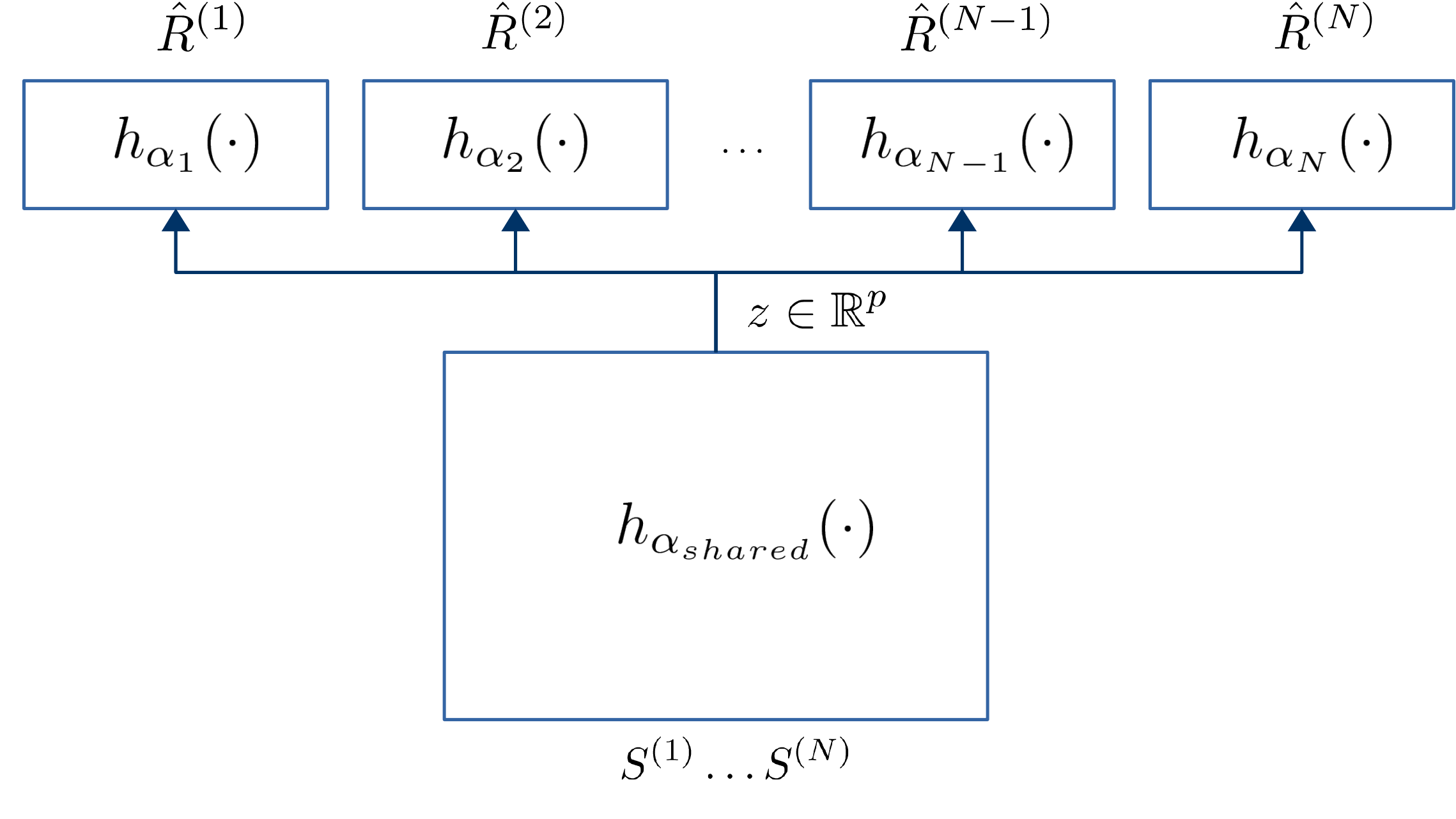}\caption[Multi-headed network for reward regression on $N$ tasks.]{Multi-headed network for reward regression on $N$ tasks.}\label{fig:nn}
\end{figure}

At this aim, we design a multi-headed network (Fig. \ref{fig:nn}) where
the states $\{S^{(i)}\}_{i=1}^N$ collected during the training of the $N$ tasks are the network input and each head outputs $\hat{R}^{(i)}=h_{\alpha_i}(h_{\alpha_{shared}}(S^{(i)}))$ are the estimate of the reward signals $R^{(i)}$ of the $i$-th task. 
The network has some  shared  layers (with parameters $\alpha_{shared}$) followed by $N$ separate heads (with parameters $\alpha_1, \dots \alpha_N$). 
	The output of the shared layers is the latent variable $z=h_{\alpha_{shared}}(s)$, with $z \in \mathbb{R}^p$ and $s \in \{S^{(i)}\}_{i=1}^N \subseteq \mathbb{R}^n$. The network is designed so as to bottleneck the output of the shared layers and obtain a shared minimal representation across tasks with lower dimension with respect to the entire state space ($p<n$). 
The shared layers and the heads of the network can be convolutional  or shallow neural networks, according to whether the  inputs $\{S^{(i)}\}_{i=1}^N$ are images or not.

The idea underlying the structure of such a network is that the shared layers should be able to learn what is important for predicting the reward function, regardless of the specific task. The latent variable $z=h_{\alpha_{shared}}(s)$ should then represent that portion of the state responsible for experiencing the rewards. Moreover, the multi-headed structure prevents
overfitting and allows better generalization~\cite{cabi2017intentional}.
The network training is formulated as a regression problem on each head.

\subsection{Exploration via maximum-entropy bonus over the state latent representation }
\label{explo}
A family of strategies for speeding up state space exploration requires to add a bonus $\mathcal{B}(s)$\cite{tang2017exploration, bellemare2016unifying, fu2017ex2, houthooft2016vime, pathak2017curiosity} to the reward function $R_{new}(s) = R(s) + \gamma \mathcal{B}(s)$\footnote{$\gamma$ is a scaling factor for properly weighting  the bonus with respect to the original reward.}.
The goal of the  bonus is to drive the learning algorithm when no or poor rewards $R(s)$ are provided. A possible choice for $\mathcal{B}(s)$ is the quantity $-log(p(s))$~\cite{fu2017ex2}, which, when using Policy Gradient for training the policy $\pi$ parametrized in $\theta$, makes the overall objective equal to:
\begin{equation}
\max_{\theta}{E_{p_{\theta}(a,s)}[R(s)] + H(p_{\theta}(s))},
\label{max}
\end{equation}
with $H(p_{\theta}(s))=E_{p_{\theta}(s)}[-log(p_{\theta}(s))]$ the entropy over the state distribution $p(s)$.
The  policy $\pi$ resulting from solving Eq. \ref{max}  maximizes both the original reward  and the entropy over the state space density distribution $p(s)$. Maximizing the entropy of a distribution entails making the distribution as uniform as possible. The uniform distribution on a finite space  is in fact the maximum entropy distribution among all continuous distributions that are supposed to be in the same space. This results in encouraging the  policy to explore states  not visited yet.

 \begin{figure}[t!]
 	\centering
 	\includegraphics[width=0.7\columnwidth]{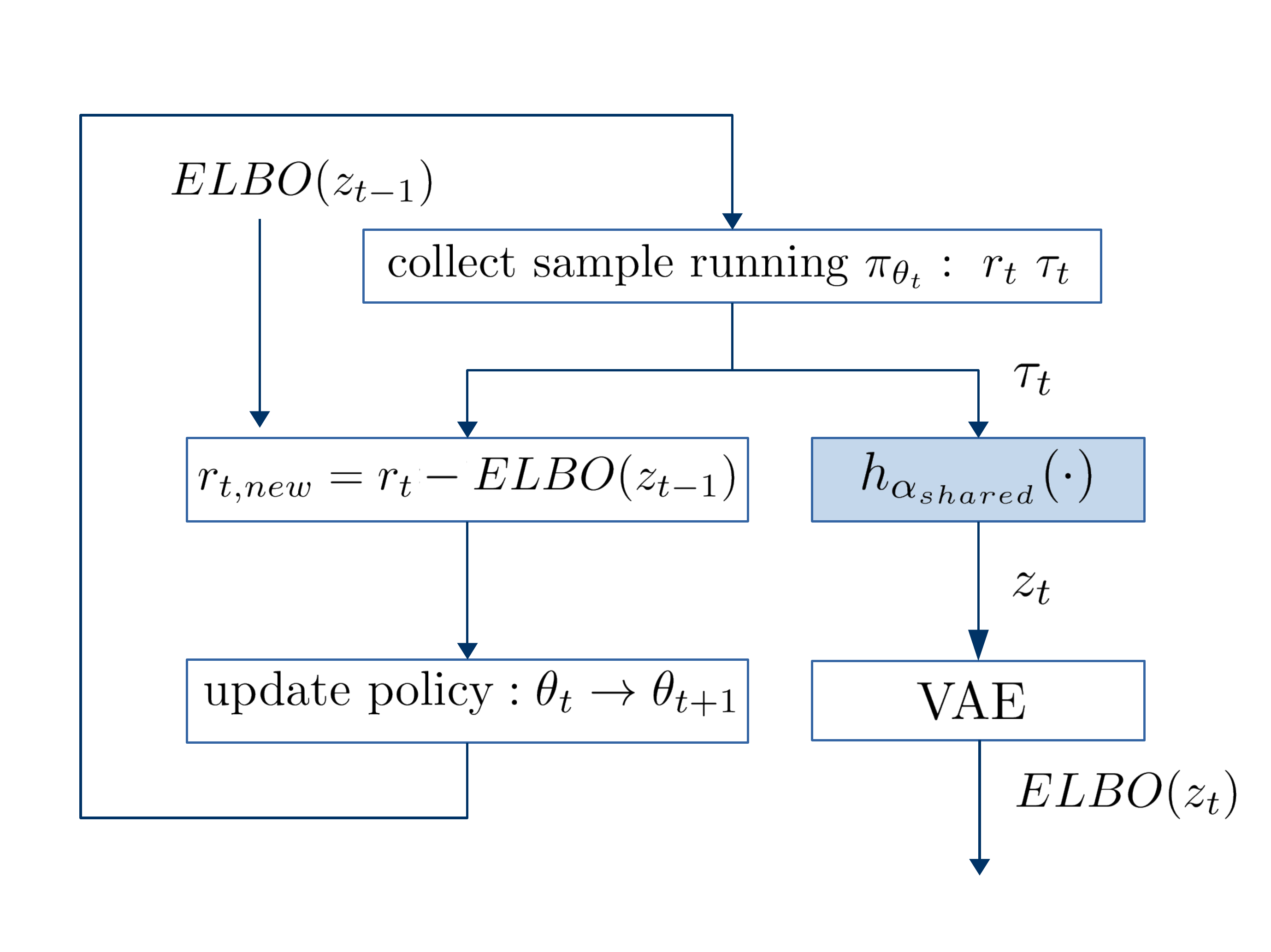}
 	\caption[Maximum-entropy bonus exploration on a learned latent representation $z$.]{Maximum-entropy bonus exploration on a learned latent representation $z$. Note that $h_{\alpha_{shared}}(\cdot)$ was trained offline, using the data collected during the solution of prior tasks. During the maximum-entropy bonus exploration algorithm its parameters are kept fixed.\label{fig:explo-alg}}
 \end{figure}
 
Even if reasonable, the choice of visiting all the possible states is not efficient when dealing with state space with high dimensionality or  rewards that can be experienced only in a small subregion of the space. For this reason, the augmented reward function  we use for exploration is given by:
 \begin{equation}
 R_{new}(s) = R(s) -  \gamma log(p(z)),
 \label{rews}
 \end{equation}
 where $z=h_{\alpha_{shared}}(s) \in \mathbb{R}^p$ ($p<n$)  is the latent representation learned from prior tasks  (Paragraph \ref{multihead}).  Maximizing the entropy only over $z$ is much more efficient because it represents the only portion of the state responsible for the rewards and has lower dimensionality with respect  to the states $s \in \mathbb{R}^n$~\cite{hazan2018provably}.
 
 In order to estimate the quantity $-log(p(z))$, we use a Variational AutoEncoder (VAE) \cite{kingma2013auto}. As shown in Appendix \ref{A1}, the final loss function after the training of the VAE can be expressed as:
 \begin{equation}
 \label{elbo}
 L(\psi, \phi)=-ELBO \simeq -log(p(z)),
 \end{equation}
 therefore providing a good approximation of $-log(p(z))$.
We can therefore augment our reward function using the  $-ELBO$:
 \begin{equation}
 R_{new}(s) = R(s) - \gamma ELBO = R(s) - \gamma log(p(z)).
 \end{equation}

 The final exploration algorithm is summarized in Algorithm \ref{explo-alg} and graphically represented in Fig. \ref{fig:explo-alg}.
 \begin{algorithm}[t!]
 \footnotesize
 	\caption{Maximum-entropy bonus exploration}
 	\label{explo-alg}
 	\begin{algorithmic}
 		\STATE Initialize the policy $\pi_{\theta_0}$, the VAE encoder and decoder;
 		\STATE Use one \textit{on policy} PG algorithm (TRPO~\cite{Schulman15} in our case);
 		\FOR{$ t=1, \dots, t_{max}$ }
 		\STATE Collect data  $(\tau_t, r_t)$ running $\pi_{\theta_t}$ ;
 		\STATE Compute $r_{t,new} = r_t -  ELBO(z_{t-1})$;
 		\STATE Update policy parameters according the algorithm in use: $\theta_t \rightarrow \theta_{t+1}$
 		\STATE  Compute the latent representation $z_{t}=h_{\alpha_{shared}}(\tau_{t})$;
 		\STATE Train VAE on $z_t$.
 		\ENDFOR
 	\end{algorithmic}
 \end{algorithm}
 

\section{Results}
\label{results}
The proposed algorithm has been tested on the object-pusher environment (Fig. \ref{fig:pusher-env}), fully described in Appendix \ref{A2}.
The reward function is sparse in that it consists of
 the Euclidean distance between the green object and the target and is different from 0 only when the object is sufficiently close to the target, i.e.  in the circle with center $g$ and radius $\delta$. The value $\delta$ regulates the sparsity of the rewards.

\begin{figure}[t!]
		\centering
		\includegraphics[width=0.8\columnwidth]{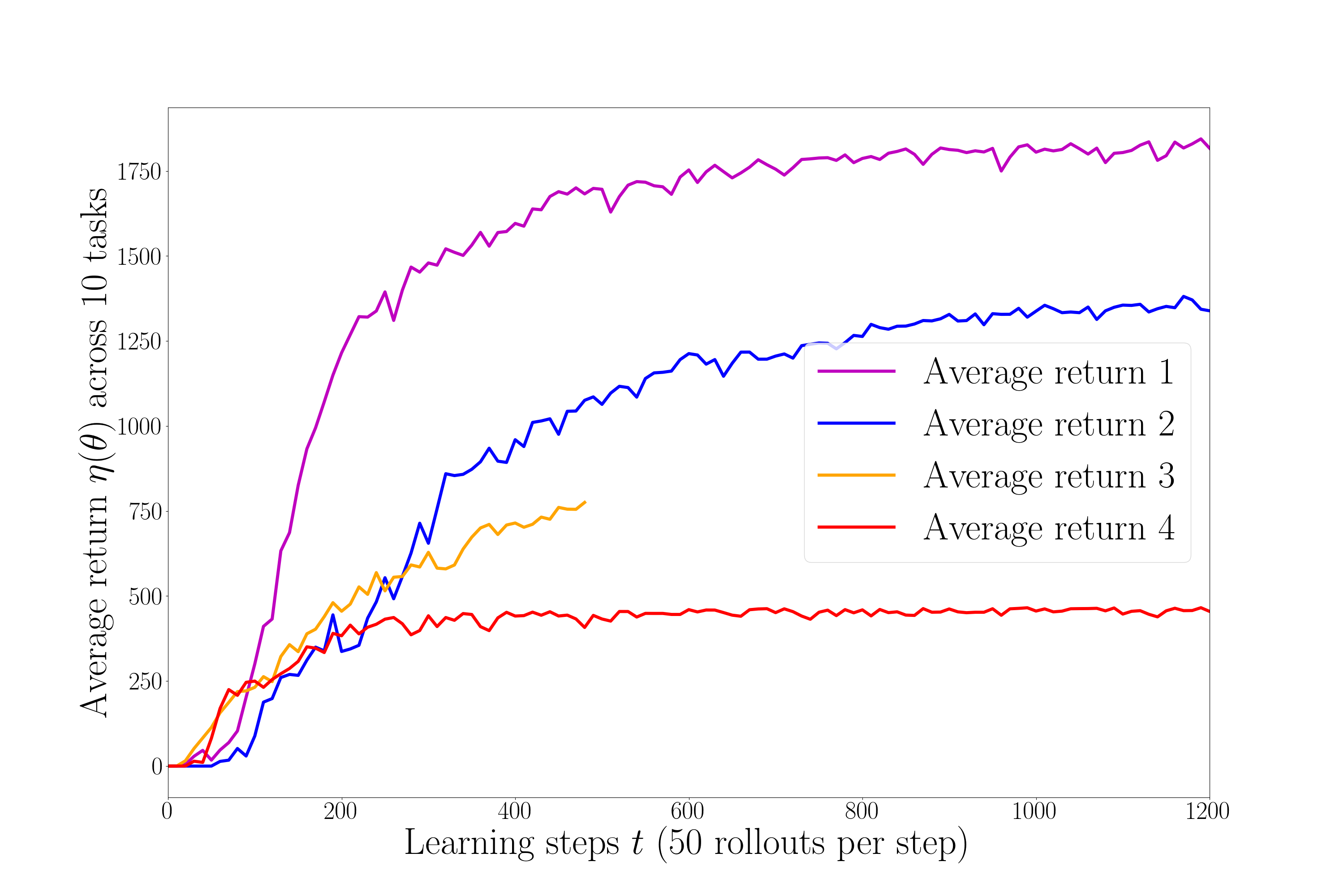}
		\caption[Average return $\eta(\theta)$ of our approach and some basic baselines.]{Average return $\eta(\theta)$ of our approach and some basic baselines when training 10 new tasks. In particular, we compare 1) our approach (Average return 1, in magenta); 2)
		TRPO  with maximum-entropy bonus over the \textit{entire state} $s$ (Average return 2, in blue); 3) TRPO with maximum-entropy bonus over the \textit{action state} (Average return 3, in orange) and 4)
		 TRPO with no exploration bonus in the reward function (Average return 4, in red). \label{fig:baselines}}
	\end{figure}

We train three policies $\pi_1, \pi_2, \pi_3$ in order to maximize the following rewards:
	\begin{equation}
	R_1=R_{pusher}(o_0) - \gamma log(p_{\theta}(o_0)),
		\end{equation}
		\begin{equation}
	R_2=R_{pusher}(o_0) - \gamma log(p_{\theta}(z)),
	\end{equation}
		\begin{equation}
	R_3=R_{pusher}(o_0) - \gamma log(p_{\theta}(s)).
	\end{equation}
	\begin{figure}[t!]
		\centering
		\includegraphics[width=0.8\columnwidth]{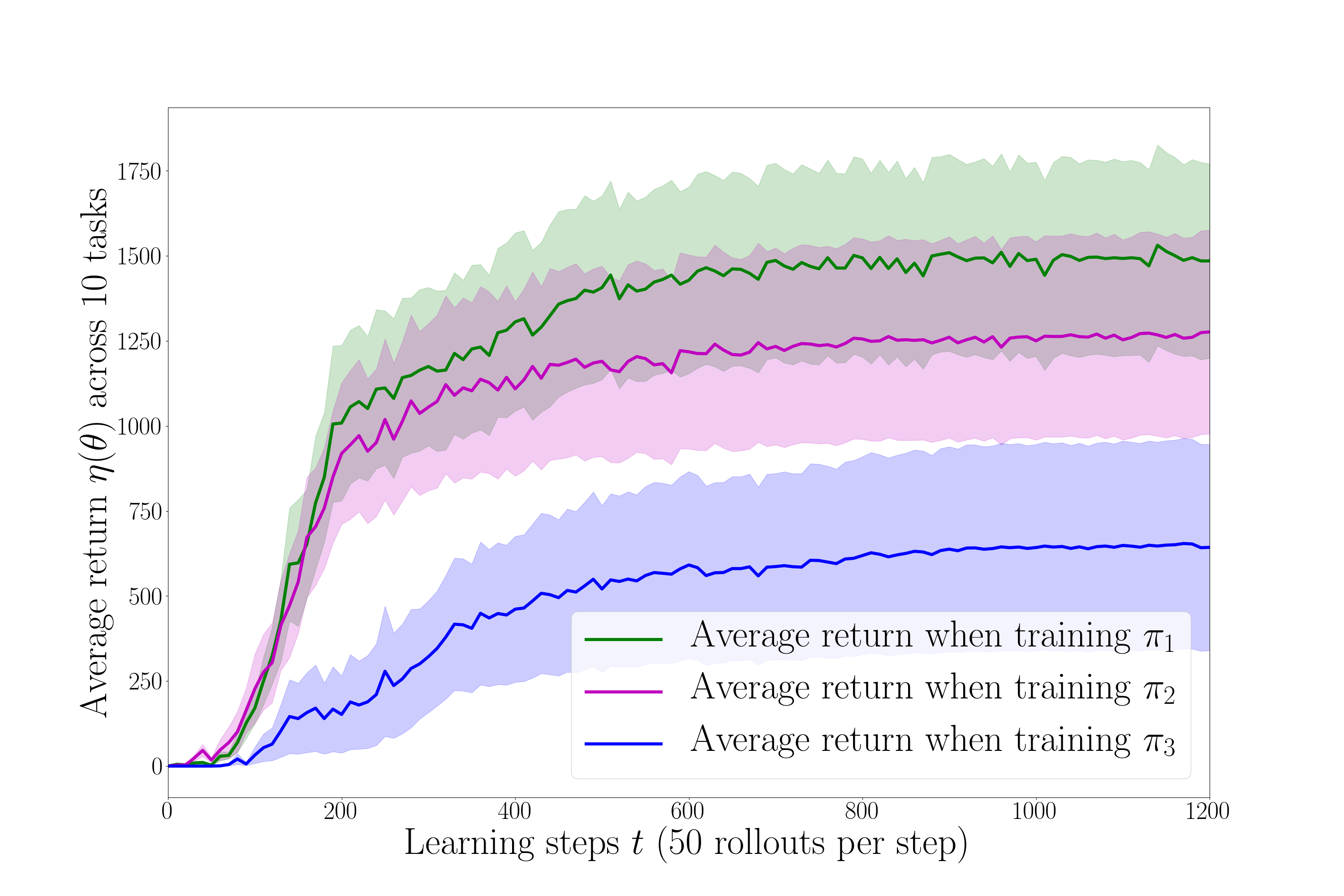}
		\caption[Average returns $\eta(\theta)$ with different reward bonus.]{ Average returns $\eta(\theta)$ with different reward bonus when training 10 new tasks. \label{fig:returns1}}
	\end{figure}
	Fig. \ref{fig:returns1} reports the average returns $\eta(\theta)$ (defined in Eq. \ref{goal}) obtained when training $\pi_1, \pi_2$ and $\pi_3$ on 10 new tasks sampled from the distribution $P_T$ We consider the training curve obtained with reward $R_1$ as an oracle, because in this case the task is solved by maximizing the entropy of the \textit{exact position} of the object of interest $o_0$, i.e. the portion of the state responsible for the rewards. The training curve obtained  when training $\pi_3$ is treated as a baseline, since the policy is required to maximize the entropy over the entire state $s$. The plot shows that  maximizing $R_2=R_{pusher}(o_0) - \gamma log(p_{\theta}(z))$ (i.e.  our approach) leads to performance almost as good as using the oracle and considerably better than running the baseline. This is at the same time a proof of the effectiveness of our learned latent representation $z$ and of our maximum-entropy bonus exploration strategy.
 We finally compare our approach with some basic baselines in Fig. \ref{fig:baselines}. This experiment confirms that the approach presented in this work outperform some baselines commonly used for addressing exploration. Further experiments are provided in Appendix \ref{A3}.

\section{Conclusions}
\label{conclusions}
In this work, we proposed a novel  exploration algorithm that leads effectively the training  of the desired task by using information learned from the solution of similar prior tasks. In particular, we proposed a multi-headed network (Section \ref{multihead}) able to encode in its shared layers the portion of state space  effective at predicting the reward in the task family of interest. This information was then encoded into an exploration algorithm based on entropy maximization (Section \ref{explo}). The experiments we carried out (Section \ref{results}) showed that the proposed method  leads to a faster solution of new tasks sampled from the same task distribution. The promising results encourage us to extend this work by including processing from raw pixels and tests on more complex tasks.
\bibliography{example_paper}

\begin{thebibliography}{23}
\providecommand{\natexlab}[1]{#1}
\providecommand{\url}[1]{\texttt{#1}}
\expandafter\ifx\csname urlstyle\endcsname\relax
  \providecommand{\doi}[1]{doi: #1}\else
  \providecommand{\doi}{doi: \begingroup \urlstyle{rm}\Url}\fi

\bibitem[Bellemare et~al.(2016)Bellemare, Srinivasan, Ostrovski, Schaul,
  Saxton, and Munos]{bellemare2016unifying}
Bellemare, M., Srinivasan, S., Ostrovski, G., Schaul, T., Saxton, D., and
  Munos, R.
\newblock Unifying count-based exploration and intrinsic motivation.
\newblock In \emph{Advances in Neural Information Processing Systems}, pp.\
  1471--1479, 2016.

\bibitem[Cabi et~al.(2017)Cabi, Colmenarejo, Hoffman, Denil, Wang, and
  De~Freitas]{cabi2017intentional}
Cabi, S., Colmenarejo, S.~G., Hoffman, M.~W., Denil, M., Wang, Z., and
  De~Freitas, N.
\newblock The intentional unintentional agent: Learning to solve many
  continuous control tasks simultaneously.
\newblock \emph{arXiv preprint arXiv:1707.03300}, 2017.

\bibitem[Chentanez et~al.(2005)Chentanez, Barto, and
  Singh]{chentanez2005intrinsically}
Chentanez, N., Barto, A.~G., and Singh, S.~P.
\newblock Intrinsically motivated reinforcement learning.
\newblock In \emph{Advances in neural information processing systems}, pp.\
  1281--1288, 2005.

\bibitem[Clavera et~al.(2018)Clavera, Rothfuss, Schulman, Fujita, Asfour, and
  Abbeel]{clavera2018model}
Clavera, I., Rothfuss, J., Schulman, J., Fujita, Y., Asfour, T., and Abbeel, P.
\newblock Model-based reinforcement learning via meta-policy optimization.
\newblock \emph{arXiv preprint arXiv:1809.05214}, 2018.

\bibitem[Doersch(2016)]{doersch2016tutorial}
Doersch, C.
\newblock Tutorial on variational autoencoders.
\newblock \emph{arXiv preprint arXiv:1606.05908}, 2016.

\bibitem[Finn et~al.(2017)Finn, Abbeel, and Levine]{finn2017model}
Finn, C., Abbeel, P., and Levine, S.
\newblock Model-agnostic meta-learning for fast adaptation of deep networks.
\newblock \emph{arXiv preprint arXiv:1703.03400}, 2017.

\bibitem[Fortunato et~al.(2017)Fortunato, Azar, Piot, Menick, Osband, Graves,
  Mnih, Munos, Hassabis, Pietquin, et~al.]{fortunato2017noisy}
Fortunato, M., Azar, M.~G., Piot, B., Menick, J., Osband, I., Graves, A., Mnih,
  V., Munos, R., Hassabis, D., Pietquin, O., et~al.
\newblock Noisy networks for exploration.
\newblock \emph{arXiv preprint arXiv:1706.10295}, 2017.

\bibitem[Fu et~al.(2017)Fu, Co-Reyes, and Levine]{fu2017ex2}
Fu, J., Co-Reyes, J., and Levine, S.
\newblock Ex2: Exploration with exemplar models for deep reinforcement
  learning.
\newblock In \emph{Advances in Neural Information Processing Systems}, pp.\
  2577--2587, 2017.

\bibitem[Gupta et~al.(2018)Gupta, Mendonca, Liu, Abbeel, and
  Levine]{gupta2018meta}
Gupta, A., Mendonca, R., Liu, Y., Abbeel, P., and Levine, S.
\newblock Meta-reinforcement learning of structured exploration strategies.
\newblock \emph{arXiv preprint arXiv:1802.07245}, 2018.

\bibitem[Hazan et~al.(2018)Hazan, Kakade, Singh, and
  Van~Soest]{hazan2018provably}
Hazan, E., Kakade, S.~M., Singh, K., and Van~Soest, A.
\newblock Provably efficient maximum entropy exploration.
\newblock \emph{arXiv preprint arXiv:1812.02690}, 2018.

\bibitem[Houthooft et~al.(2016)Houthooft, Chen, Duan, Schulman, De~Turck, and
  Abbeel]{houthooft2016vime}
Houthooft, R., Chen, X., Duan, Y., Schulman, J., De~Turck, F., and Abbeel, P.
\newblock Vime: Variational information maximizing exploration.
\newblock In \emph{Advances in Neural Information Processing Systems}, pp.\
  1109--1117, 2016.

\bibitem[Kingma \& Welling(2013)Kingma and Welling]{kingma2013auto}
Kingma, D.~P. and Welling, M.
\newblock Auto-encoding variational bayes.
\newblock \emph{arXiv preprint arXiv:1312.6114}, 2013.

\bibitem[Nair et~al.(2018)Nair, McGrew, Andrychowicz, Zaremba, and
  Abbeel]{nair2018overcoming}
Nair, A., McGrew, B., Andrychowicz, M., Zaremba, W., and Abbeel, P.
\newblock Overcoming exploration in reinforcement learning with demonstrations.
\newblock In \emph{2018 IEEE International Conference on Robotics and
  Automation (ICRA)}, pp.\  6292--6299. IEEE, 2018.

\bibitem[Pathak et~al.(2017)Pathak, Agrawal, Efros, and
  Darrell]{pathak2017curiosity}
Pathak, D., Agrawal, P., Efros, A.~A., and Darrell, T.
\newblock Curiosity-driven exploration by self-supervised prediction.
\newblock In \emph{International Conference on Machine Learning (ICML)}, volume
  2017, 2017.

\bibitem[Plappert et~al.(2017)Plappert, Houthooft, Dhariwal, Sidor, Chen, Chen,
  Asfour, Abbeel, and Andrychowicz]{plappert2017parameter}
Plappert, M., Houthooft, R., Dhariwal, P., Sidor, S., Chen, R.~Y., Chen, X.,
  Asfour, T., Abbeel, P., and Andrychowicz, M.
\newblock Parameter space noise for exploration.
\newblock \emph{arXiv preprint arXiv:1706.01905}, 2017.

\bibitem[Rajeswaran et~al.(2018)Rajeswaran, Kumar, Gupta, Vezzani, Schulman,
  Todorov, and Levine]{rss}
Rajeswaran, A., Kumar, V., Gupta, A., Vezzani, G., Schulman, J., Todorov, E.,
  and Levine, S.
\newblock Learning complex dexterous manipulation with deep reinforcement
  learning and demonstrations.
\newblock \emph{Robotics: Science and Systems (RSS)}, 2018.

\bibitem[Schulman et~al.(2015)Schulman, Levine, Abbeel, Jordan, and
  Moritz]{Schulman15}
Schulman, J., Levine, S., Abbeel, P., Jordan, M., and Moritz, P.
\newblock Trust region policy optimization.
\newblock In \emph{International Conference on Machine Learning}, pp.\
  1889--1897, 2015.

\bibitem[Sendonaris \& Dulac-Arnold(2017)Sendonaris and
  Dulac-Arnold]{sendonaris2017learning}
Sendonaris, A. and Dulac-Arnold, C.~G.
\newblock Learning from demonstrations for real world reinforcement learning.
\newblock \emph{arXiv preprint arXiv:1704.03732}, 2017.

\bibitem[Stadie et~al.(2015)Stadie, Levine, and
  Abbeel]{stadie2015incentivizing}
Stadie, B.~C., Levine, S., and Abbeel, P.
\newblock Incentivizing exploration in reinforcement learning with deep
  predictive models.
\newblock \emph{arXiv preprint arXiv:1507.00814}, 2015.

\bibitem[Tang et~al.(2017)Tang, Houthooft, Foote, Stooke, Chen, Duan, Schulman,
  DeTurck, and Abbeel]{tang2017exploration}
Tang, H., Houthooft, R., Foote, D., Stooke, A., Chen, O.~X., Duan, Y.,
  Schulman, J., DeTurck, F., and Abbeel, P.
\newblock \#{E}xploration: A study of count-based exploration for deep
  reinforcement learning.
\newblock In \emph{Advances in Neural Information Processing Systems}, pp.\
  2753--2762, 2017.

\bibitem[Taylor et~al.(2011)Taylor, Suay, and Chernova]{taylor2011integrating}
Taylor, M.~E., Suay, H.~B., and Chernova, S.
\newblock Integrating reinforcement learning with human demonstrations of
  varying ability.
\newblock In \emph{The 10th International Conference on Autonomous Agents and
  Multiagent Systems-Volume 2}, pp.\  617--624. International Foundation for
  Autonomous Agents and Multiagent Systems, 2011.

\bibitem[Van~Hoof et~al.(2015)Van~Hoof, Hermans, Neumann, and
  Peters]{van2015learning}
Van~Hoof, H., Hermans, T., Neumann, G., and Peters, J.
\newblock Learning robot in-hand manipulation with tactile features.
\newblock In \emph{Humanoid Robots (Humanoids), 2015 IEEE-RAS 15th
  International Conference on}, pp.\  121--127. IEEE, 2015.

\bibitem[Vecer{\'\i}k et~al.(2017)Vecer{\'\i}k, Hester, Scholz, Wang, Pietquin,
  Piot, Heess, Roth{\"o}rl, Lampe, and Riedmiller]{vecerik2017leveraging}
Vecer{\'\i}k, M., Hester, T., Scholz, J., Wang, F., Pietquin, O., Piot, B.,
  Heess, N., Roth{\"o}rl, T., Lampe, T., and Riedmiller, M.~A.
\newblock Leveraging demonstrations for deep reinforcement learning on robotics
  problems with sparse rewards.
\newblock \emph{CoRR, abs/1707.08817}, 2017.

\end{thebibliography}
\bibliographystyle{icml2019}

\newpage

\appendix
\section{Entropy computation with VAE}
\label{A1}
In this Section, we explain how the VAE is used in order to estimate $-log(p(z))$.
The structure of the VAE  is the following: 

 \begin{figure}[h!]
 	\includegraphics[width=\columnwidth]{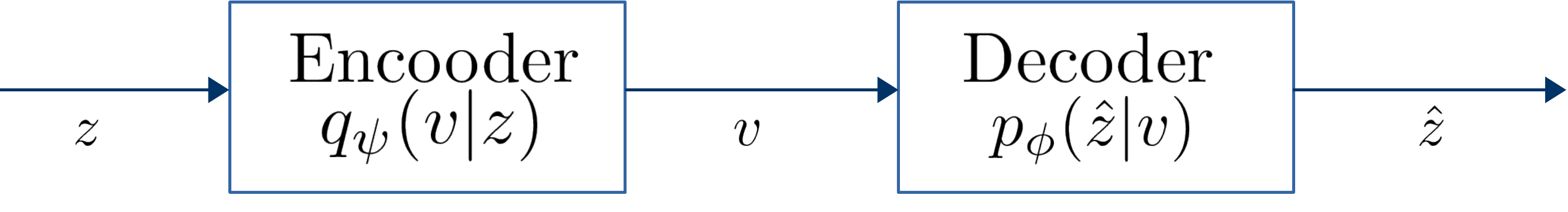}\caption[Variational Autoencoder.]{Variational Autoencoder.}\label{fig:VAE}
 \end{figure}
 \begin{itemize}
 	\item The input of the encoder $q_{\psi}(v| \cdot)$ is the latent representation $z=h_{\alpha_{shared}}(s)$. The procedure to learn $h_{\alpha_{shared}}(\cdot)$ has been presented in Paragraph \ref{multihead}.
 	\item $v$ is the latent variable reconstructed by the VAE, i.e. the output of the encoder and input of the decoder . This quantity is not relevant for our formulation since we are not interested in dimensionality reduction. We just mention it for the sake of completeness.
  	\item The output $\hat{z}$ of the decoder  $p_{\phi}( \cdot | v)$  is the reconstruction of the input ${z}$.

 \end{itemize}
 The loss minimized during VAE training~\cite{doersch2016tutorial} is given by:
 \begin{equation}
 \label{loss-vae}
 L(\psi, \phi)= - E_{q_{\psi}(v|z)} [log(p_{\phi}(\hat{z}|v)] + D_{KL}(q_{\phi}(v|z) || p(v)),
 \end{equation}
 where $D_{KL}$ is the Kullback-Leibler divergence between the encoder distribution $q_{\phi}(v|z)$ and  the distribution $p(v)$. The first term of Eq. \ref{loss-vae} is the \textit{reconstruction loss}, or expected negative log-likelihood. This term encourages the decoder to learn to reconstruct the data. The second term is a regularizer that measures the information lost when using $q_{\phi}(v|z) $ to represent $p(v)$. In variational autoencoders $p(v)$ is chosen to be a standard Normal distribution $p(v)=N(0,1)$. 
 
 The VAE loss function in Eq. \ref{loss-vae} can be proved to be equal to the negative evidence lower bound (ELBO)~\cite{kingma2013auto}, that is defined as:
 \begin{equation}
 -\text{ELBO}=-log(p(z)) + D_{KL} (q_{\phi}(v|z) || p(v|z)).
 \end{equation}
$p(v|z)$ cannot be computed analytically, because it describes the values of $v$ that are likely to  provide a sample similar to $z$ using the decoder. The KL divergence imposes the distribution $q_{\phi}(v|z)$ to be close to $p(v|z)$.
If  we use an arbitrarily
 high-capacity model for $q_{\phi}(v|z)$, we can assume that - at the end of the training - $q_{\phi}(v|z)$ actually match $ p(v|z)$ and the KL-divergence term is close to zero~\cite{kingma2013auto}. As a result, the final loss function after the training of the VAE can be expressed as:
 \begin{equation}
 L(\psi, \phi)=-\text{ELBO} \simeq -log(p(z)),
 \end{equation}
 therefore providing a good approximation of $-log(p(z))$.

\section{2D pusher environment}
\label{A2}
\begin{figure}
	\centering
	\framebox{\parbox{0.3\linewidth}{\includegraphics[width=0.3\columnwidth]{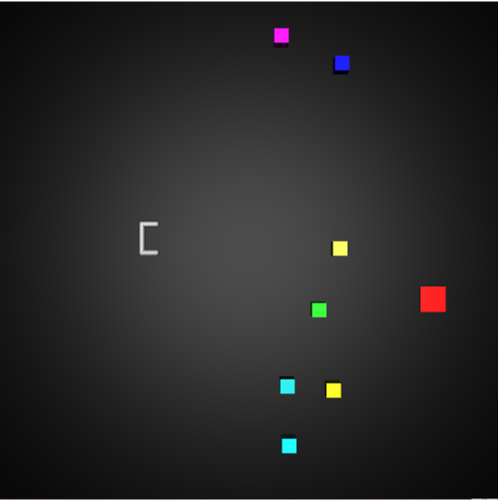}}}\caption[2D object-pusher environment.]{2D object-pusher environment. The task consists of moving the green object towards the target, represented with a red square. \label{fig:pusher-env}}
\end{figure}

Here is the full description of the environment used during the tests (Fig. \ref{fig:pusher-env}).
\begin{itemize}
    \item The goal is to push the green object (identified as object no. $0$) towards the red target.
	\item The environment state includes the objects ($o_0, \dots, o_6$), pusher ($p$) and target\footnote{The target position is constant during time.} ($g$) 2D positions:
	\begin{equation}
	s=[o_0, o_1, \dots, o_6, p, g].	
	\end{equation}
	The 2D space of the environment is finite and continuous.
	\item The action space is 2D and continuous, allowing the pusher to move forward - backward and laterally.
	\item The reward function is given by:
	\begin{equation}
	\label{rewards}
	R = R_{pusher}(o_0)=1\{d(o_0, g)< \delta\} d(o_0, g)^2,
	\end{equation}
	where $d(o_0, g)$ is the Euclidean distance between the green object and the target. The reward is then  differnt from zero only when the object is sufficiently close to the target, i.e.  in the circle with center $g$ and radius $\delta$. The value $\delta$ regulates the sparsity of the rewards. In our experiments with considered $\delta=0.1$ in a space of dimension $2\times2$.
	\item Different tasks of this environment differ in the initial object positions and target position.
 \end{itemize}
 
 \section{Analysis on the exploration trajectories}
 \label{A3}
 In order to better analyze the performance of the proposed exploration algorithm, we train three policies $\pi_{1}, \pi_2, \pi_3$ in order to maximize the rewards:
	\begin{equation}
	R_1=-log(p_{\theta}(o_0)),
	\end{equation}
	\begin{equation}
	R_2=-log(p_{\theta}(z)),
	\end{equation}
	\begin{equation}
	R_3=-log(p_{\theta}(s)).
	\end{equation}
	The three  policies are therefore trained in order to make respectively the distribution of the trajectories  of the 1) \textit{object of interest position} $o_0$, 2) the \textit{latent representation $z$}  and 3) the \textit{entire state} $s$ as uniform as possible.
	\begin{figure}[h!]
		\centering
		\includegraphics[width=0.8\columnwidth]{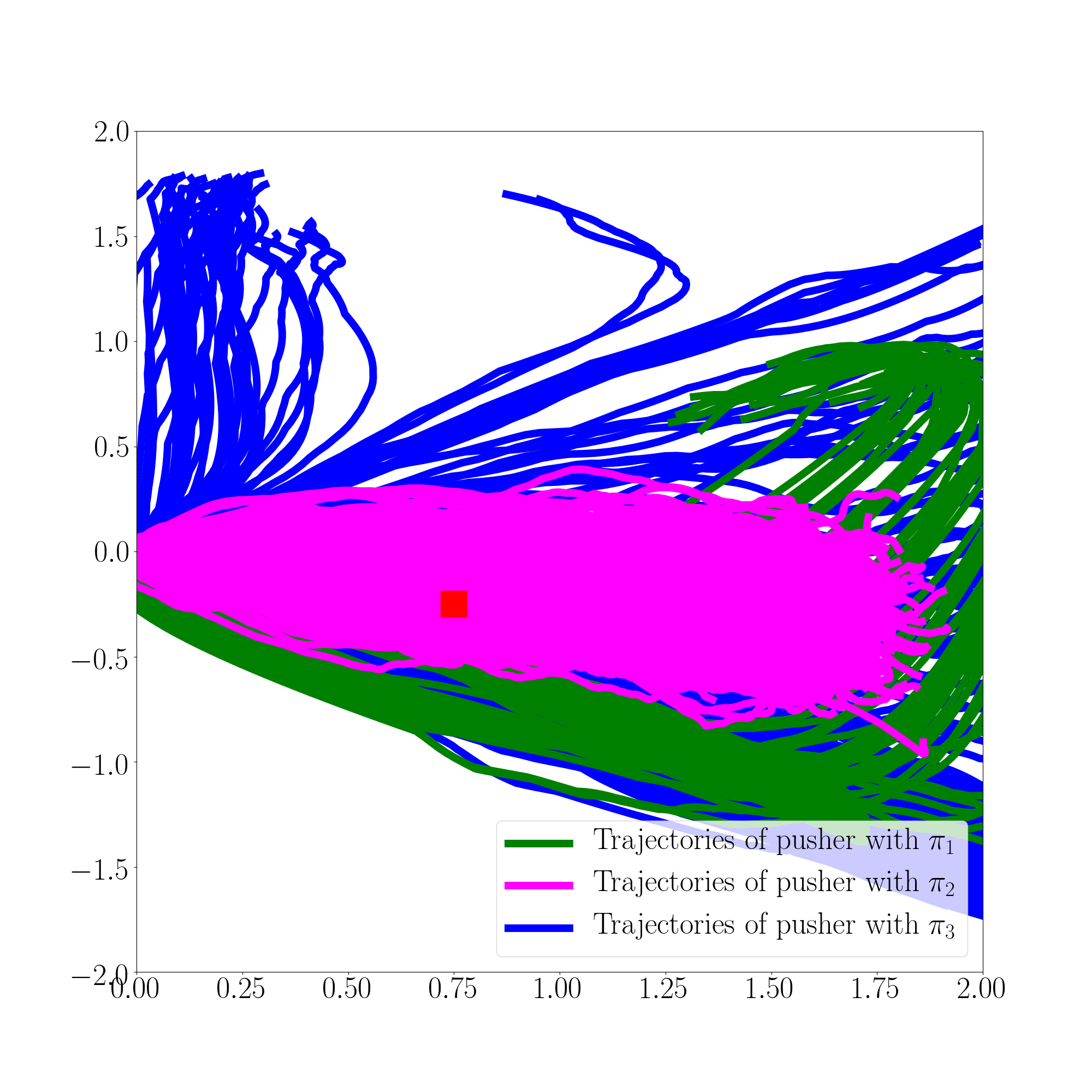}
		\caption[100 pusher trajectories obtained when running each trained policy $\pi$, $\pi_2$, $\pi_3$.]{100 pusher trajectories obtained when running the trained policy $\pi$, $\pi_2$, $\pi_3$. The initial position object the object of interest $o_0$ is represented with a red square. \label{fig:traj-pusher}}
	\end{figure}
		\begin{figure}[h!]
			\centering
			\includegraphics[width=0.8\columnwidth]{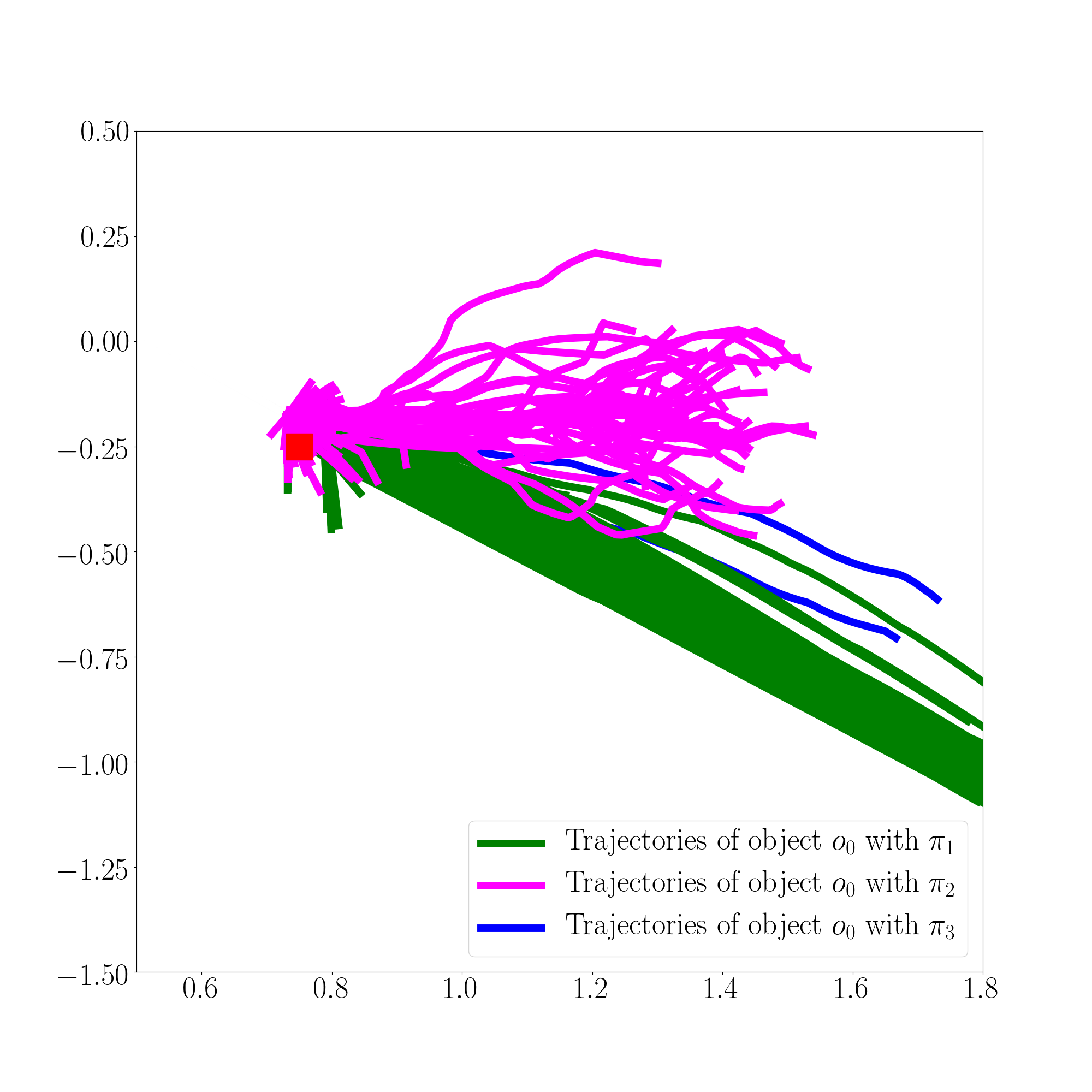}
			\caption[Object trajectories obtained when running  100 times each trained policy $\pi$, $\pi_2$, $\pi_3$.]{Object trajectories obtained when running  100 times each trained policy $\pi$, $\pi_2$, $\pi_3$. The number of trajectories shown is less than 100 for each policy because not all the policy executions lead to movements of the object of interest. \label{fig:traj-object}}
		\end{figure}
	The policies are trained by using Algorithm \ref{explo-alg}, with $R_{new}=R_1, R_2$ and $R_3$ and, when rewards $R_1$ and $R_3$ are used, without using the learned latent representation $z$ but directly feeding the VAE with, respectively, $o_0$ and $ s$ to estimate $-log(p_{\theta}(o_0))$ and $-log(p_{\theta}(s))$.  Figs. \ref{fig:traj-pusher} and \ref{fig:traj-object} report respectively the trajectories followed by the \textit{pusher} and the \textit{object of interest $o_0$} when running the three trained policies. When the policy is asked to maximize only the object of interest trajectories, the pusher (green trajectories in Fig. \ref{fig:traj-pusher})  focuses the efforts in moving  towards the object of interest (whose initial position is represented with  a red square). The consequent object trajectories are shown in green in Fig. \ref{fig:traj-object}.  Analogous trajectories both for the pusher and the object are generated by $\pi_2$, obtained  by maximizing $R_2=-log(p_{\theta}(z))$ (in magenta in \ref{fig:traj-pusher} and \ref{fig:traj-object}), meaning that our latent representation correctly encodes the position of the object of interest. Instead, in the third case the pusher explores the entire state and the object is rarely pushed (blue trajectorie in \ref{fig:traj-pusher} and \ref{fig:traj-object}).
\end{document}